\documentclass[a4paper,conference]{IEEEtran}
\ifCLASSINFOpdf
\else
\fi
\usepackage{graphicx}
\usepackage{xcolor}\usepackage{booktabs}
\usepackage{amsfonts}
\usepackage{amsmath}
\usepackage{bm}
\begin{document}
%
\title{Towards cumulative race time regression in sports: I3D ConvNet transfer learning in ultra-distance running events}

\author{\IEEEauthorblockN{David Freire-Obreg\'on, Javier Lorenzo-Navarro, Oliverio J. Santana,\\ Daniel Hern\'andez-Sosa and Modesto Castrill\'on-Santana}
\IEEEauthorblockA{SIANI\\
Universidad de Las Palmas de Gran Canaria\\
Spain\\
Email: david.freire@ulpgc.es}
}


%


\maketitle

\begin{abstract}
Predicting an athlete's performance based on short footage is highly challenging. Performance prediction requires high domain knowledge and enough evidence to infer an appropriate quality assessment. Sports pundits can often infer this kind of information in real-time. In this paper, we propose regressing an ultra-distance runner cumulative race time (CRT), i.e., the time the runner has been in action since the race start, by using only a few seconds of footage as input. We modified the I3D ConvNet backbone slightly and trained a newly added regressor for that purpose. We use appropriate pre-processing of the visual input to enable transfer learning from a specific runner. We show that the resulting neural network can provide a remarkable performance for short input footage: 18 minutes and a half mean absolute error in estimating the CRT for runners who have been in action from 8 to 20 hours. Our methodology has several favorable properties: it does not require a human expert to provide any insight, it can be used at any moment during the race by just observing a runner, and it can inform the race staff about a runner at any given time.
\end{abstract}


%
\IEEEpeerreviewmaketitle

\begin{figure*}[bt] 
\begin{minipage}{1\linewidth}
    \centering
    \includegraphics[scale=0.55]{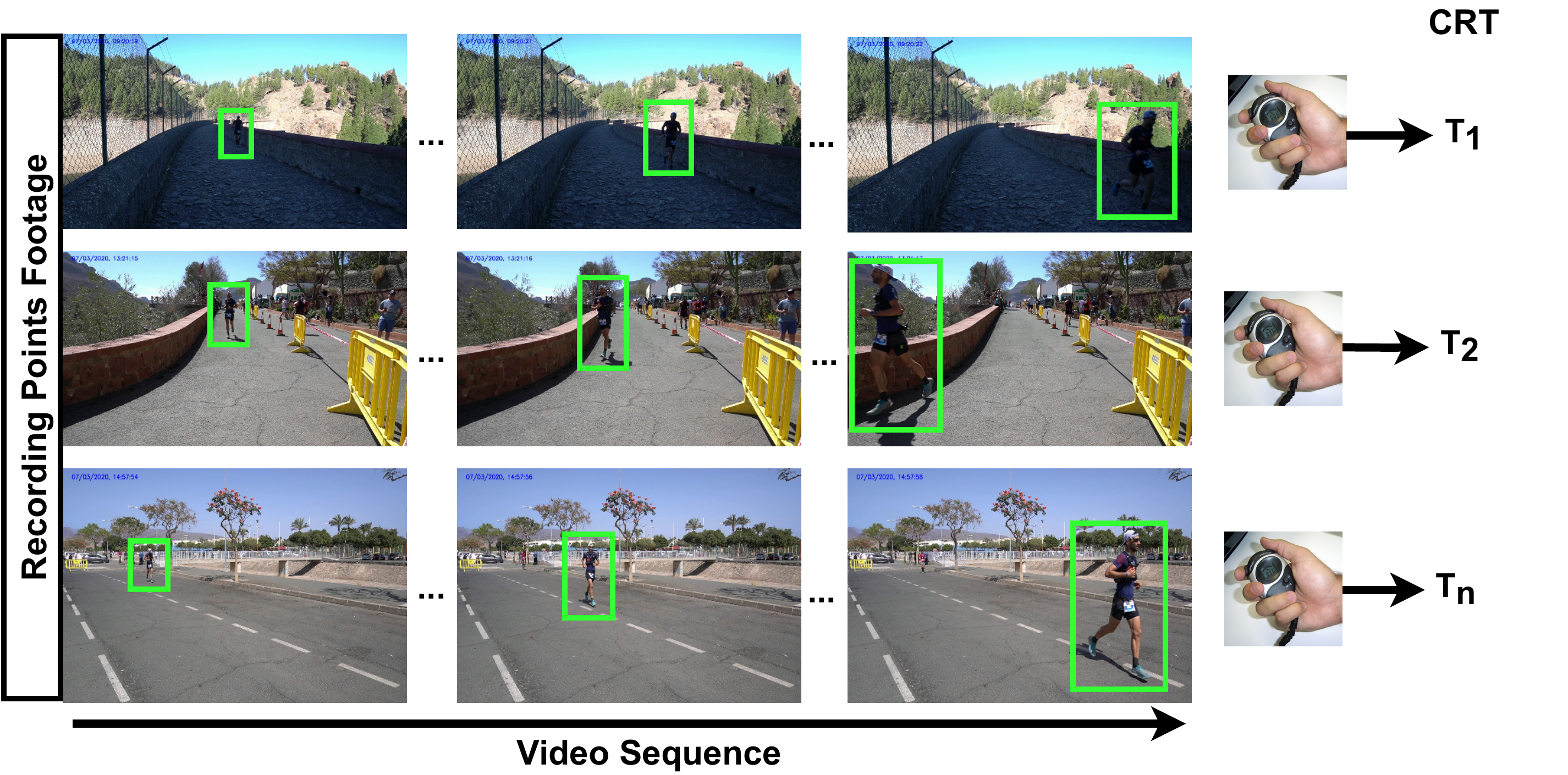}
      \end{minipage}
    \caption{\textbf{Collage of a runner's footage in set of recording points.} Green containers correspond to body detections. We used an I3D ConvNet for each footage that efficiently interprets actions to extract the runner's embeddings. Then, these embeddings are fed into a model to infer the CRT at a specific recording point. The CRT at a particular recording point $RP_{i}$ can be defined as $T_{i} = T_{1} + \sum_{j=2}^{j=i} (T_{j} - T_{r}) \; where \; r=j-1.$}
     \label{fig:intro_img}
\end{figure*}

\section{Introduction}

Our ability to point out and describe an athlete's performance relies on a solid sports understanding of his/her actions and all of the elements around him/her. Understanding any sport requires acknowledging its metrics, which is the element that modifies the scoreboard. In ultra-distance races, the runner's assessment depends on the CRT. Like cycling events, athletes pass through several specific locations where their CRTs are registered. Eventually, the first runner to reach the finish line wins the race. Precisely, recording the split time at different locations provides an insight into the runner's performance at a given partial track.


There has been significant progress in sports events detection in the last few years. Several applications have benefited from this progress, including summarization or highlight extraction \cite{He20,Gao20}. For instance, tracking a ball in individual sports -i.e., tennis \cite{Qazi15}- or in team sports -i.e., basketball \cite{Wei15}- can be connected in a later semantic associated step such as scoreboard update. In contrast, non-ball-based sports -i.e., running or cycling- have received less attention. Two fundamental challenges when analyzing videos are 1) the localization of the relevant moments in a video and 2) the classification/regression of these moments into specific categories/data. In our problem, the former is dataset-dependent, but the latter focuses on regressing the input footage. These problems are even more complicated in ultra-distance races, where highly dynamic scenes and the race span represent challenges during the acquisition process.

This work aims to overcome these issues by proposing a novel pipeline non-dependent on the acquisition instant, neither on the scenario. First, the footage input is preprocessed to focus on the runner of interest (see Figure \ref{fig:intro_img}). Then, an I3D Convolutional Neural Network (I3D) is modified to extract the most meaningful features at the end of its convolutional base. 
Additionally, these features will feed a brand new regressor for the time regression task. Our model is inspired by recent work in action recognition using I3D \cite{Joefrie19,Wang19,Dai21,Freire22}. Our second contribution introduces a complete ablation study on how single and double input streams affect regression. 

We have evaluated our model in a dataset collected to evaluate Re-ID methods in real-world scenarios. The collection contains $214$ ultra-distance runners captured at different recording points along the track. The achieved results are remarkable (up to 18,5 minutes of MAE), and they have also provided interesting insights. The first is the importance of context when considering the two 3D stream inputs and how this input outperforms every single 3D stream input. Another insight is related to the importance of contextual information for the pre-trained I3D and the limitations observed during the transfer learning. In this sense, more informative contexts are up to a $50$\% better than restrictive context (i.e., bounding boxes). Finally, we have divided the dataset observations into their timing quartiles. In this regard, all the models have shown a moderate MAE degradation, up to a $39$\% in the best model.

The paper is organized into five sections. The next section discusses some related work. Section \ref{sec:pipeline} describes the proposed pipeline. Section \ref{sec:results} reports the experimental setup and the experimental results. Finally, conclusions are drawn in Section \ref{sec:con}.

\section{Related Work}
\label{sec:sota}

Sports analysis has continually drawn the Community's attention, leading to a rapid increase in published work during the last decade. According to the semantic significance, the sports video content can be divided into four layers: raw video, object, event, and semantic layers \cite{Shih18}. The pyramid base comprises the input videos from which objects can be identified in a superior layer. In this sense, key objects represented in the video clips can be identified by extracting objects -i.e., players \cite{Tianxiao20}- and tracking objects -i.e., ball \cite{Shaobo19}, and players \cite{JungSoo20}-. The event layer represents the action of the key object. Several actions combined with scene information can generate an event label, and it represents the related action or interaction among multiple objects. Works connected to action recognition \cite{Freire22}, trajectory prediction \cite{Teranishi20} and highlight detection \cite{Gao20} fit this layer scope. The top layer is the semantic layer, representing the footage semantic summarization \cite{Cioppa18}. Since our purpose is regressing a CRT from athlete footage, we aim to obtain an estimated outcome from the athlete's actions, a value that defines his performance.

In this regard, several sports collections have been gathered from international competitions events to boost research on automatic sports quality assessment in the recent past. Some of the latest datasets are the MTL-AQA dataset (diving)~\cite{Parmar19}, UNLV AQA-7 dataset (diving, gymnastic vaulting, skiing, snowboarding, and trampoline)~\cite{Parmar19wacv} and Fis-V dataset (skating)~\cite{Xu20}. All these datasets are collected in an indoor and not-occlusive environment or, in the case of the UNLV AQA-7 dataset (snowboarding/skiing), in a quiet environment: black sky (night) and white ground (snow). Moreover, the collections mentioned above show professional athletes. Our work considers an ultra-distance race collection with a set of variations in terms of lighting conditions, backgrounds, occlusions, accessories, and athlete's proficiency diversity.

In ultra-distance races, each runner is provided with a small RFID tag worn on the runner's shoe, as a wristband, or embedded in the runner's bib. RFID readers are placed throughout the course and are used to track the runner's progress \cite{Chokchai18}. Some runners cheat despite these measures by taking shortcuts \cite{Rosie80}. Consequently, the primary research carried out around this sport is re-identification. The Community has developed different proposals such as mathematical models \cite{Goodrich21}, invasive devices similar to RFID \cite{Lingjia19}, and non-invasive models based on the runner's appearance \cite{Klontz13}, or his bib number \cite{Ben-Ami12-bmvc, Carrascosa20}.

\begin{figure*}[t]  
\begin{minipage}[t]{1\linewidth}
    \centering
    \includegraphics[scale=0.7]{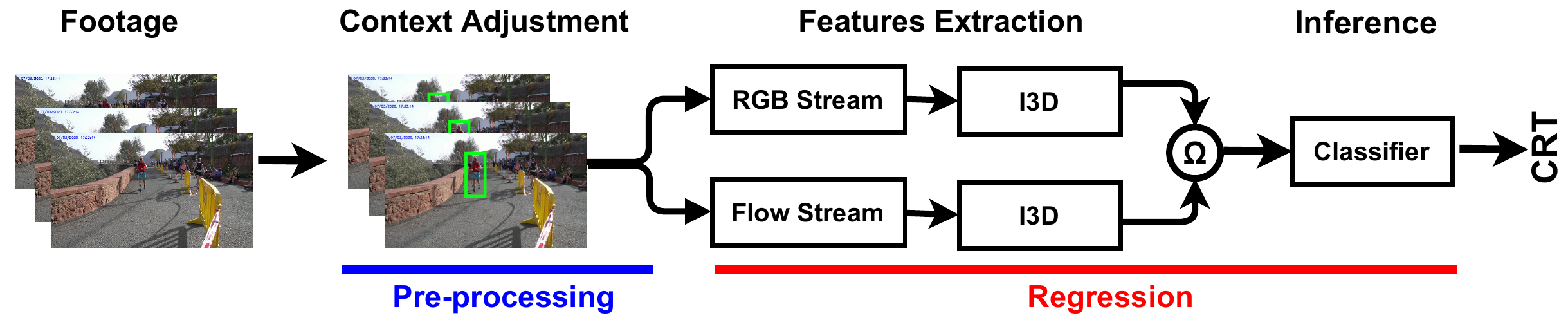}
      \end{minipage}
    \caption{\textbf{The proposed pipeline for the CRT regression problem.} The devised process comprises two main parts: the footage pre-processing block and the regression block. In the former, the tracker assists by helping to neutralize the runner background activity (SB or BB). The latter implies the generation of two streams of data (RGB and Flow) that feed two I3D ConvNets whose outputs can be either added or concatenated ($\Omega$). The resulting tensor acts as an input to the regressor, ending the regression.
     \label{fig:pipeline}}
\end{figure*}

In the last two decades, gait analysis has been explored differently. The human pose representation plays a crucial role in evaluating a performed action. Lei et al. have identified three primary pose representations for the action quality assessment~\cite{Lei19}. The challenge relies on finding robust features from pose sequences and establishing a method to measure the similarity of pose features~\cite{Wnuk10}. Second, the skeleton-based representations encode the relationship between joints and body parts~\cite{Freire20}. However, the estimated skeleton data can often be noisy in realistic scenes due to occlusions or changing lighting conditions~\cite{Carissimi18}, especially in an ultra-distance race in wild conditions. 

More recently, deep learning methods have been applied to analyze gait. Some authors, like Parmar, suggest that in this representation approach, convolutional neural networks (CNN) can be combined with recurrent neural networks (RNN) due to the temporal dimension provided by the video input~\cite{Parmar17}. Another typical network used for sports in terms of quality assessment is the 3D convolution network (C3D). This deep neural network can learn spatio-temporal features that define action quality~\cite{Tran14}. The Residual Network (ResNet) has played a key role in the action recognition task in the last few years. Thus, Hara proposed the 3D ResNets improving the recognition performance by increasing the ResNet layers depth \cite{Hara18}. Both C3D and 3D ResNets operate on single 3D stream input. Precisely, our work can be framed in the deep learning methods for assessing the athlete's action quality suggested by Lei et al. We make use of the I3D ConvNet, which has been used for human action recognition (HAR) in the past~\cite{ Freire22,Carreira17}. Unlike C3D and 3D ResNets, the I3D ConvNet operates on two 3D stream inputs.

In summary, the work presented in this paper regresses the race participant's CRT considering an I3D network. This regression provides meaningful semantic information connected to the runner's performance. Since the considered dataset provides a scenario in the wild, we evaluate our pipeline after pre-processing the raw video sequence as input to remove context noisy elements.

\section{Proposed architecture}
\label{sec:pipeline}
As can be seen in Figure \ref{fig:pipeline}, this work proposes a sequential pipeline divided into two major blocks. The I3D ConvNet requires a footage input for inference, and it processes two streams: RGB and optical flow. Consequently, any object (athlete, race staff, cars) not of interest must be removed from the scene. Precisely, the first block pre-preprocess the raw input data to focus on the runner of interest. The output of this block can comprise the runner bounding box (BB) or the runner with still background (SB), see Figure \ref{fig:extract}. Figure \ref{fig:pipeline} also shows how the first block feeds the I3D convolutional base. Then, each pre-processed footage is divided into the two streams mentioned earlier to feed a pre-trained I3D ConvNet. Finally, the extracted features feed a regressor to infer the CRT.

\subsection{Footage pre-processing}
\label{sec:footpress}

A few years ago, Deep SORT \cite{Wojke17} showed significant advances in tracking performance. Then, the SiamRPN+ \cite{zhang2019deeper} lead to remarkable robustness by taking advantage of a deeper backbone like ResNet. Additionally, Freire et al. \cite{Freire22} have combined both Deep SORT and SiamRPN+ to increase the tracking precision when illumination changes or partial occlusions happen in HAR in some sports collections. Recently, ByteTrack has outperformed DeepSort as multi-object tracking by achieving consistent improvements \cite{zhang2021bytetrack}. This tracker uses the YOLOX detector \cite{ge2021yolox} as the backbone, and a COCO pre-trained model \cite{Lin14} as the initialized weights. This tracker has exhibited remarkable occlusion robustness and outperformed the previously described trackers. Therefore, we have used ByteTrack to track the runner of interest.

As we mentioned before, a context adjustment pre-processing can generate the scenarios considered in our experiments, defined as BB and SB. Given a runner $i$ bounding box area $BB_i(t, RP)$ at a given time $t \in [0,T]$ and in a recording point $RP \in [0,P]$, the new pre-processed footage $F'_i[RP]$ can be formally denoted as follows:

\begin{equation}
\label{eq:contextremoval}
F'_i[RP]=BB_{i}(t, RP) \cup \tau(RP)
\end{equation}

\begin{figure}[bt] 
\begin{minipage}{1\linewidth}
    \centering
    \includegraphics[scale=0.47]{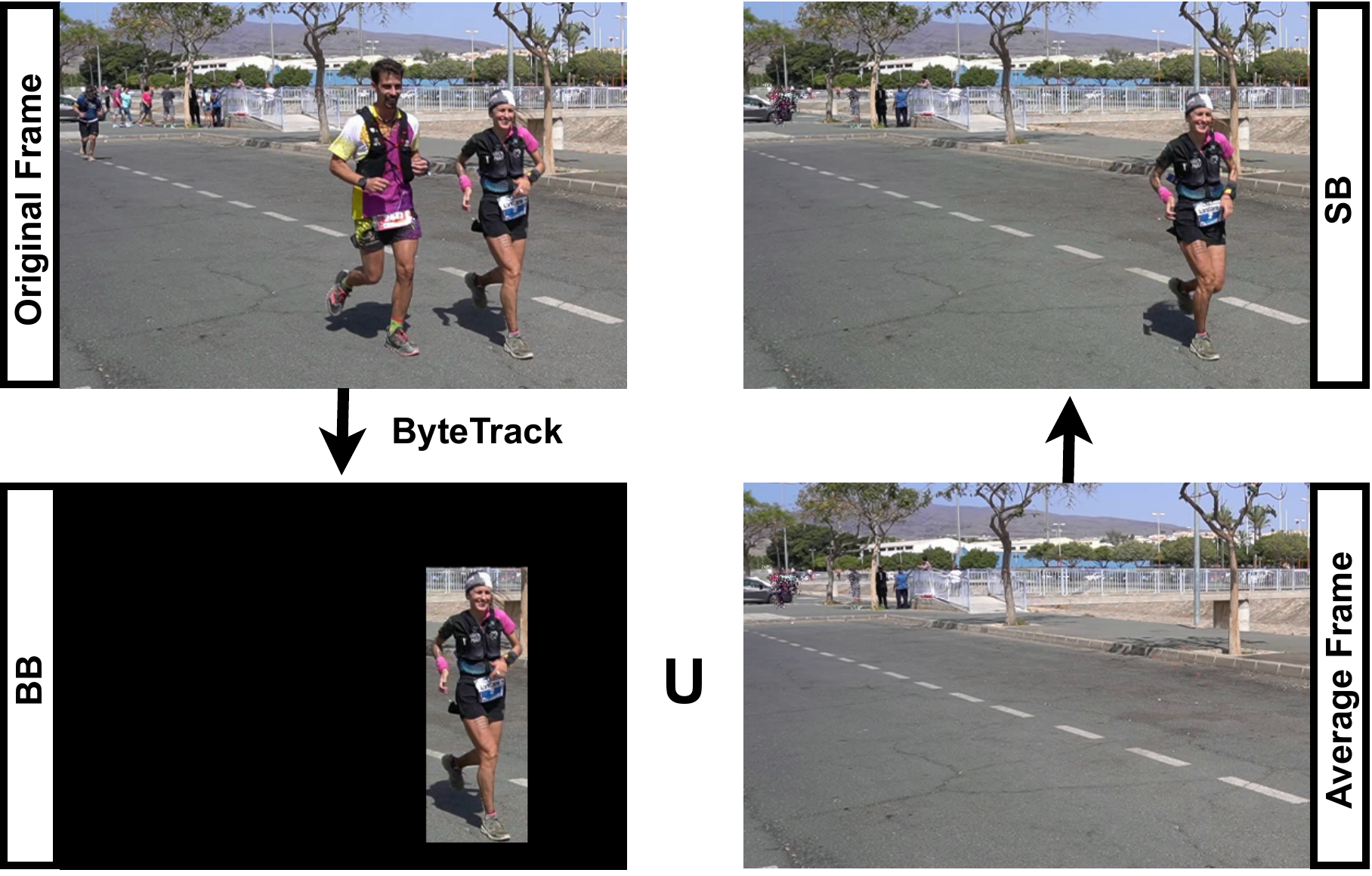}
      \end{minipage}
    \caption{\textbf{Context adjustment.} Since the I3D ConvNet processes every action in a scene, we have cleaned the context to focus on the runner of interest.}
     \label{fig:extract}
\end{figure}

Where $\tau(RP)$ can stand for the footage average frame or a neutral frame to generate the SB and BB scenario, respectively (see Figure \ref{fig:extract}).

\subsection{Runners Features Extraction}
\label{sec:i3d}

\begin{figure}[bt] 
\begin{minipage}{1\linewidth}
    \centering
    \includegraphics[scale=0.5]{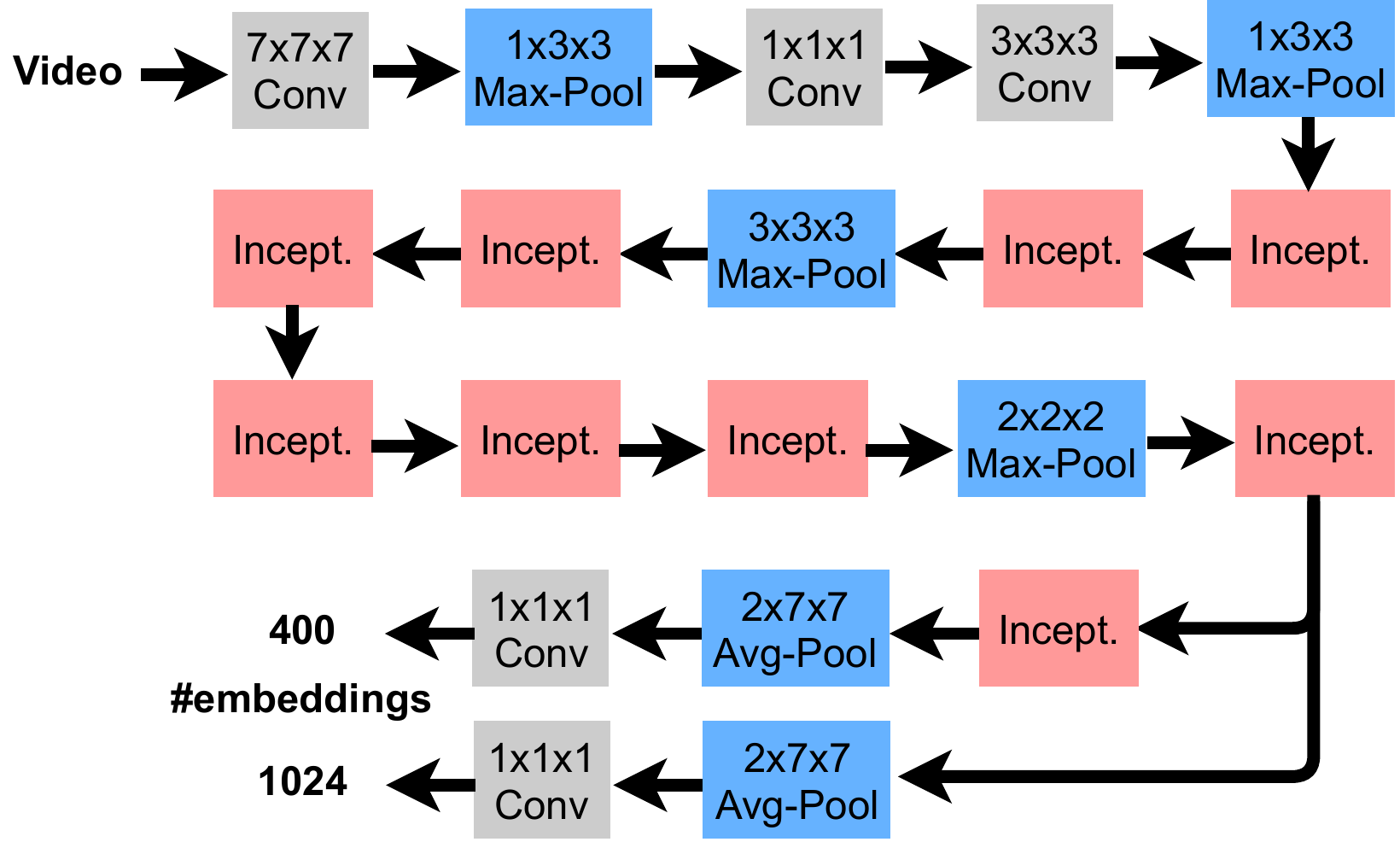}
      \end{minipage}
    \caption{\textbf{The Inflated Inception-V1 architecture.} Three main blocks can be appreciated in the diagram: convolutional layers, max/average-pool layers, and inception blocks \cite{Carreira17}. The considered outputs for our experiments can be appreciated at the bottom of the diagram. The end-to-end architecture provides 400 embeddings, whereas we also process the information from an upper layer (1024 embeddings). Both streams, RGB and optical flow are processed through this architecture.}
     \label{fig:model}
\end{figure}

A few years ago, Carreira and Zisserman proposed the Inflated 3D  Convolutional Neural Network (I3D ConvNet) based on a two-stream network \cite{Carreira17} (see Figure \ref{fig:model}). This deep neural network applies a two-stream structure for RGB and optical flow to the Inception-v1 \cite{Szegedy15} module along with 3D CNNs. In the Inception module, the input is processed by several parallel 3D convolutional branches whose outputs are then merged back into a single tensor. Moreover, a residual connection reinjects previous representations into the downstream data flow by adding a past output tensor to a later output tensor. The residual connection aims to prevent information loss along with the data-processing flow. 

Nowadays, the I3D ConvNet is one of the most common feature extraction methods for video processing \cite{Freire22}. The approach presented in this work exploits the pre-trained model on the Kinetics dataset as a backbone model \cite{Carreira17}. In our case, we have used the Kinetics \cite{Kay17} that includes $400$ action categories. Consequently, the I3D ConvNet acts as a feature extractor to encode the network input into a $400$ vector feature representation, see Figure \ref{fig:model}. However, since we are not looking for HAR but more fine-action recognition, we also considered the features obtained from a previous layer to provide more insights into the athlete's movement. In our work, we have removed the last inception block by straightly performing an average pooling and a max-pooling to reduce dimensionality to the 1024 most meaningful features. 

 In the I3D ConvNet, optical flow and RGB streams are processed separately through the architecture shown in Figure \ref{fig:model}. Then, the streams 400 output logits are added before applying a softmax layer for HAR. In the experiments conducted in Section \ref{sec:results} we have considered these 400 logits, but also the streams concatenation, devising an 800 vector feature representation. The same happens with the feature representation vectors from the previous inception block. In this case, the resulting embeddings contain 1024 and 2048 elements for the sum and the concatenation, respectively.

A trainable regressor was built on top of this network. In this sense, four different classifiers were tested during the conducted experiments: Linear regression, Random Forest, Gradient Boosting, SVM, and a Multi-Layer Perceptron.

\section{Experiments and results}
\label{sec:results}

This section is divided into two subsections related to the setup and results of the designed experiments. The first subsection describes the dataset adopted for the proposed problem and technical details, such as the method to tackle the problem. The achieved results are summarized in the second subsection.

\subsection{Experimental setup}

We design our evaluation procedure to address the following questions. Is the presented approach good enough to infer a CRT? What is the role of input context in facilitating an accurate regression process? How well does the model infer different timing percentiles? Finally, we validate our design choices with ablation studies. 

\begin{figure}[bt] 
\begin{minipage}{1\linewidth}
    \centering
    \includegraphics[scale=0.5]{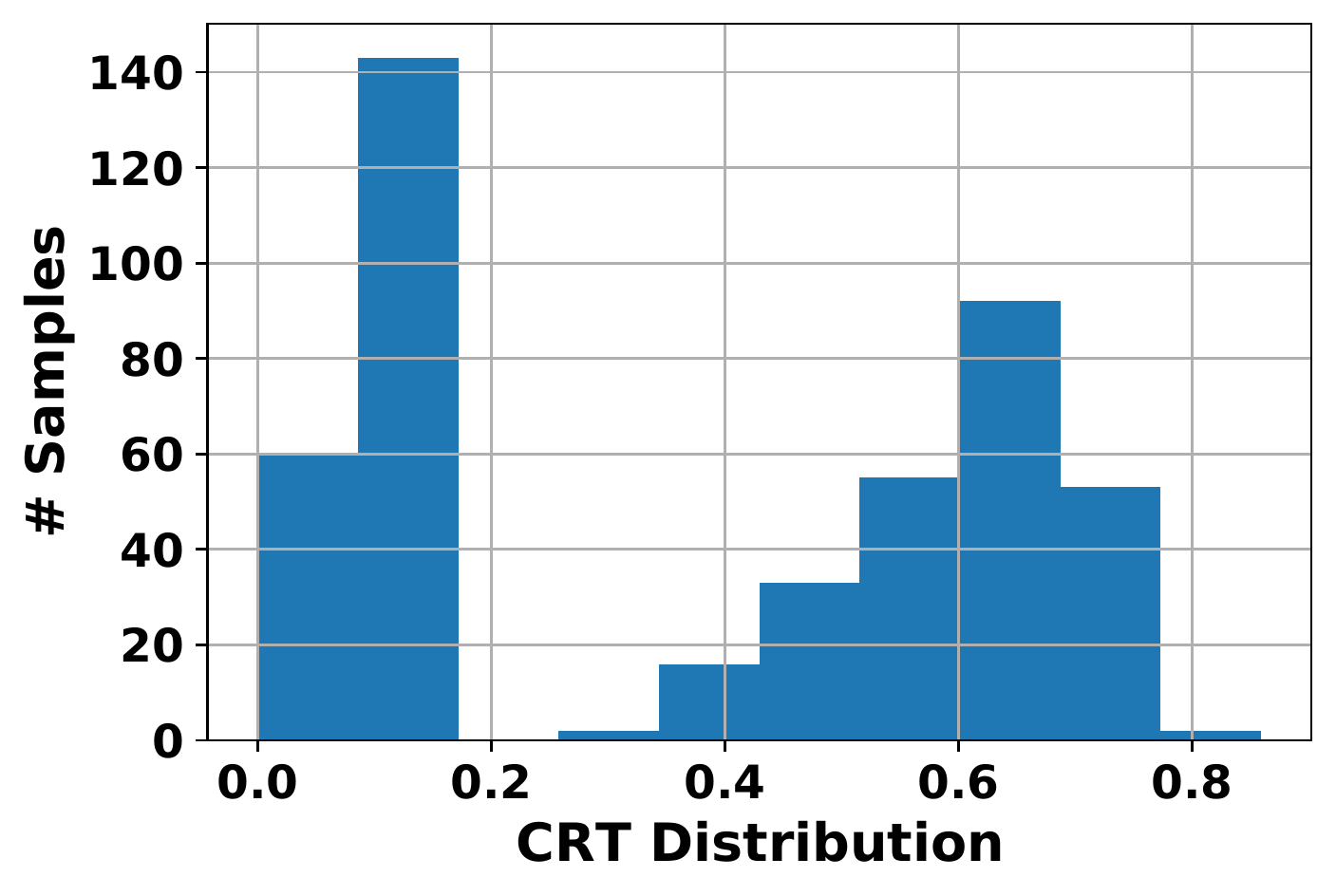}
      \end{minipage}
    \caption{\textbf{CRT distribution.} The considered dataset is balanced in terms of CRT distribution. Half of the runner's samples perform relatively well, whereas the rest perform average or worse.}
     \label{fig:sampledist}
\end{figure}

\textbf{Dataset}. We have partially used the dataset published by Penate et al.~\cite{Penate20-prl}. The mentioned dataset was collected during Transgrancanaria (TGC) 2020, an ultra-distance running competition. 
This collection comprised up to six running distances to complete the challenge, but the annotated data covers just participants in the TGC Classic who must cover 128 kilometers in less than 30 hours. The original dataset contains annotations for almost 600 participants in six different recording points. However, just 214 of them were captured after km 84. Thus just the last three recording points are considered in the experiments below. Given the different performances among participants, the gap between leaders and last runners increases along the track, while the number of participants is progressively reduced. For each participant, seven seconds clips at 25 fps captured at each recording point are fed to the footage pre-processing block described in Section \ref{sec:pipeline}. In this sense, we have used the same fps suggested by Carreira and Zisserman \cite{Carreira17}. Finally, the results presented in this section refer to the average MAE on 20 repetitions of a 10-fold cross-validation. On average, 410 samples are selected for training, and the remaining 46 samples are used for the test (Figure \ref{fig:sampledist} shows the CRT distribution). 

\textbf{Method}. A runner $i$ observation $o_i[RP] \in \mathbb{O}$ at a recording point $RP \in [0,P]$ consists of a pre-processed footage $F'_i[RP]$ and a CRT $\phi_i[RP]$. Additionally, we have normalized the runners CRT as shown in Equation \ref{eq:norm}.

\begin{equation}
    \phi'_i[RP] = \frac{\phi_i[RP] - min(\phi_i[0])}{max(\phi_i[P])}
    \label{eq:norm}
\end{equation}

Therefore, $\phi'_i[RP] \in [0,1]$.  Our task is to find an end to end regression method that minimizes the following objective:
\begin{equation}
    min\;L(\phi'_i[RP], \psi_i[RP]) = \frac{1}{N}\sum_{j=0}^{N}(\phi'_i[RP]_j-\psi_i[RP]_j)^2
    \label{eq:minim}
\end{equation}
Where $\psi_i[RP]$ stands for the runner $i$ predicted value at a recording point $RP$ just observing the runner movements up to seven seconds.

\subsection{Results}

As we stated in Section \ref{sec:pipeline}, we have considered two different scene contexts (BB and SB). Additionally, we have considered different I3D ConvNet models by adding -400 embeddings- or concatenating -800 embeddings- the last inception block output, and adding -1024 embeddings- and concatenating -2048 embeddings- the penultimate inception block output. Furthermore, we have reported rates using the following regression methods: Linear regression (LR), Random Forest (RF), Gradient Boosting (GB), SVM, and a Multi-Layer Perceptron (MLP).

\begin{table}[!htbp]
\renewcommand{\arraystretch}{1.3}
\centering
\caption{\textbf{Mean average error (MAE) achieved by each model.} The first column shows the model configuration, whereas the rest of the columns show each regression method MAE. Lower is better.}
\label{tab:mae_exp}
\begin{tabular}{|l|c|c|c|c|c|}
\hline
\#Embedd.-Stream-Cont. &  LR & RF & GB & SVM & MLP\\
\hline
400-RGB-BB       & $0.059$ & $0.061$ & $0.058$ & $0.051$ & $0.043$\\\hline
400-Flow-BB      & $0.069$ & $0.078$ & $0.076$ & $0.055$ & $0.052$\\\hline
400-RGB+Flow-BB  & $0.060$ & $0.067$ & $0.059$ & $0.052$ & $0.040$\\\hline
800-RGB$\cup$Flow-BB  & $0.057$ & $0.061$ & $0.058$ & $0.046$ & $0.034$\\\hline
\hline
400-RGB-SB       & $0.036$ & $0.034$ & $0.030$ & $0.025$ & $0.028$\\\hline
400-Flow-SB      & $0.065$ & $0.069$ & $0.068$ & $0.055$ & $0.050$\\\hline
400-RGB+Flow-SB  & $0.041$ & $0.040$ & $0.039$ & $0.032$ & $0.027$\\\hline
800-RGB$\cup$Flow-SB  & $0.036$ & $0.036$ & $0.031$ & $0.030$ & $0.019$\\\hline
\hline
1024-RGB-BB      & $0.072$ & $0.059$ & $0.059$ & $0.056$ & $0.036$\\\hline
1024-Flow-BB     & $0.077$ & $0.069$ & $0.067$ & $0.061$ & $0.042$\\\hline
1024-RGB+Flow-BB & $0.067$ & $0.057$ & $0.055$ & $0.055$ & $0.038$\\\hline
2048-RGB$\cup$Flow-BB & $0.069$ & $0.060$ & $0.057$ & $0.054$ & $0.033$\\\hline
\hline
1024-RGB-SB      & $0.039$ & $0.034$ & $0.031$ & $0.028$ & $0.017$\\\hline
1024-Flow-SB     & $0.073$ & $0.067$ & $0.062$ & $0.058$ & $0.033$\\\hline
1024-RGB+Flow-SB & $0.041$ & $0.037$ & $0.033$ & $0.032$ & $0.018$\\\hline
2048-RGB$\cup$Flow-SB & $0.040$ & $0.037$ & $0.032$ & $0.031$ & \bm{$0.015$}\\
\hline
\end{tabular}
\end{table}

Table \ref{tab:mae_exp} shows the achieved results by each model configuration. The table is divided into four blocks, having four entries each. The first block is related to the last layer embeddings when BB context is considered, the second block to the last layer embeddings when SB context is considered, the third to the penultimate layer embeddings when BB context is considered, and the last block is referred to the penultimate layer embeddings when SB context is considered.

To validate the importance of the input stream, we have analyzed the two 3D streams input -RGB+Flow- but also each 3D stream input independently -RGB or Flow-. As can be appreciated, individual streams often perform worse than the two-stream model when the best model is considered (MLP). Carreira and Zisserman stated that both streams complement each other. In other words, 3D ConvNet learns motion features from RGB input directly in a feedforward computation. In contrast, optical flow algorithms are somehow recurrent by performing iterative optimization for the flow fields. They conclude that having a two-stream configuration is better, with one I3D ConvNet trained on RGB inputs and another on flow inputs that carry optimized flow information.

Table \ref{tab:mae_exp} also highlights the relative importance of the context. As can be appreciated, SB consistently outperforms BB context. Recently, Freire et al. have informed a similar loss-performance effect using different contexts on I3D ConvNet \cite{Freire22}. They reported a 2\% to 6\% accuracy loss in a HAR problem. In our case, the loss is significant, but we can not compare it to their work due to different metrics usage and task-related issues. For HAR, background dynamics provides more insights, but it is not useful when regressing an athlete split-time since we focus on a runner activity. They stated this accuracy drop as a data loss problem during the context adjustment in their case. However, we attribute this to the characteristics of the pre-trained I3D ConvNet. As aforementioned, this network has been trained on the Kinetics that includes $400$ action categories. It has, in some sense, learned contextual features during the training process.

We further evaluate different I3D ConvNet models based on the number of considered embeddings by the classifier. Classical I3D ConvNet computes the addition -resulting 400 and 1024 embeddings- for HAR tasks. However, the reported results suggest that concatenation -800 and 2048 embeddings- provides further insight on CRT regression. Moreover, the improvement is up to 20\% in the best-reported model on Table \ref{tab:mae_exp}. The result indicates that providing more features to the regression method enhances performance no matter the source layer.

\begin{table}[!htbp]
\renewcommand{\arraystretch}{1.3}
\centering
\caption{\textbf{Comparison of different architectures on the dataset used in the present work.} The first column shows the considered pre-trained architectures, whereas the second column shows the MAE. Lower is better.}
\label{tab:comp_exp}
\begin{tabular}{|l|c|}
\hline
Architecture & MAE\\
\hline
C3D \cite{Tran14}      &  $0.038$\\\hline
3D ResNets-D30 \cite{Hara18} & $0.036$\\\hline
3D ResNets-D50 \cite{Hara18}       &  $0.033$\\\hline
3D ResNets-D101 \cite{Hara18}      & $0.032$\\\hline
3D ResNets-D200 \cite{Hara18}       &  $0.031$\\\hline
I3D-800SB (Ours)  & $0.019$\\\hline
I3D-2048SB (Ours)  & \bm{$0.015$}\\
\hline
\end{tabular}
\end{table}

Finally, the MLP outperforms any other considered classifier. In this sense, we have performed a grid search to find the most suitable architecture. It turns out that a two-layer configuration with batch normalization reported the best result. The rest of the methods are significantly far from the MAE reported by MLP. However, SVM reports the runner-up results. Consequently, a simple classifier is good enough to select the best embeddings for inference. In terms of minutes, a 0.015 MAE is roughly 18 minutes and a half. Considering that the fastest runner was recorded after 8 hours of CRT, and the last one after 20 hours of CRT, the achieved MAE can be considered a positive outcome.

To better compare the proposed pipeline with state-of-the-art proposals, we have included in Table \ref{tab:comp_exp} the best-aforementioned results. This table summarizes the performance reported in recent literature on the mentioned dataset. The table includes three major architectures, C3D, 3D ResNets, and the I3D ConvNet. Since 3D ResNets can be configured considering different depths, we show several configurations (30, 50, 101, and 200 ResNet layers). As can be appreciated in Table \ref{tab:comp_exp}, the 3D ResNets performance difference among depths is not significant beyond 50 layers. Overall, the I3D ConvNet outperforms other considered prior architectures on this task.

\begin{figure}[bt] 
\begin{minipage}{1\linewidth}
    \centering
    \includegraphics[scale=0.5]{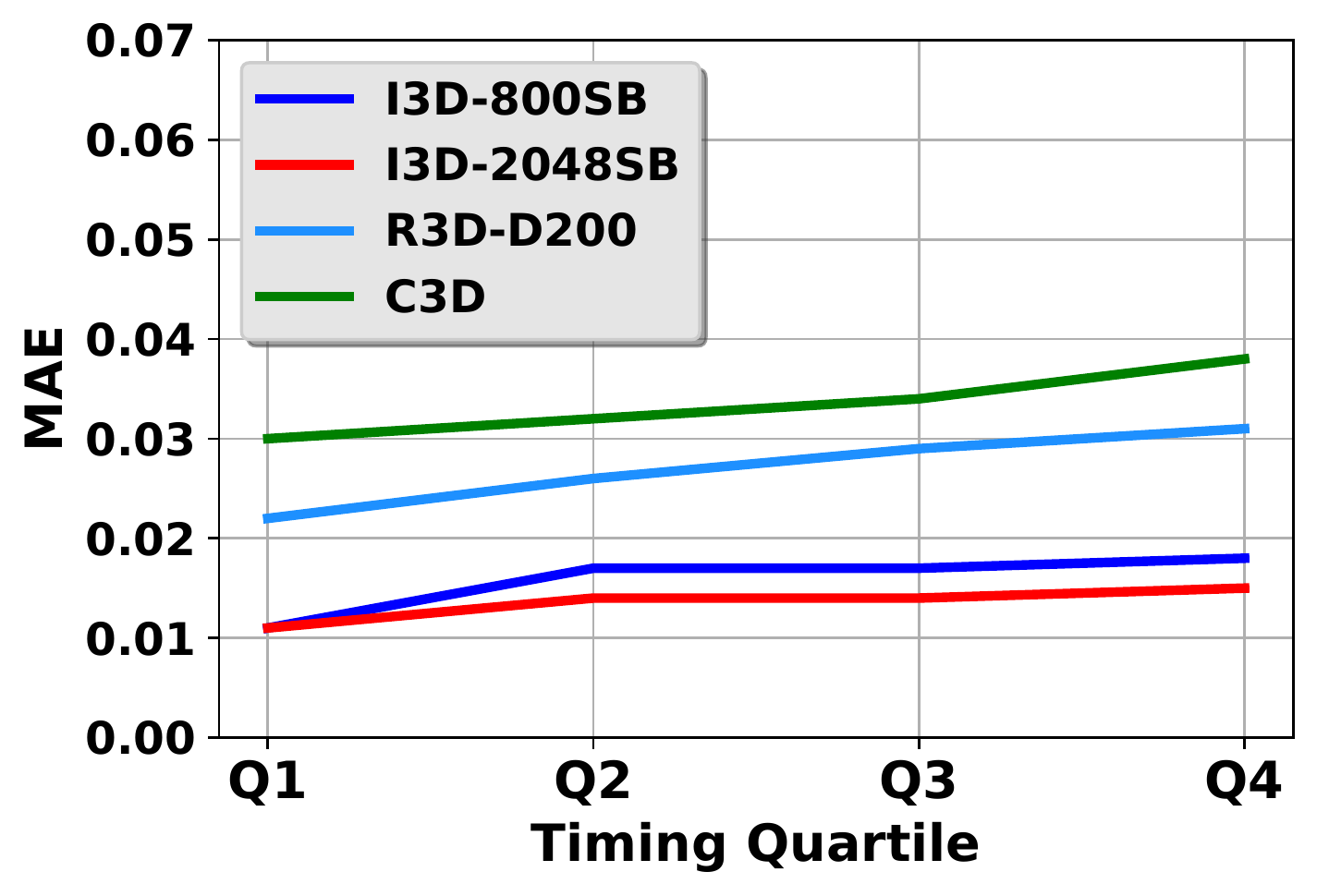}
      \end{minipage}
    \caption{\textbf{Absolute MAE increase.} This graph shows two state-of-the-art architectures and Table \ref{tab:mae_exp} best two models, concatenated embeddings from the last and penultimate layers for SB context. The degradation increases along with the timing percentile in all models.}
     \label{fig:quantile}
\end{figure}

Figure \ref{fig:quantile} analyzes the evolution of MAE for different configurations over the CRT distribution quartiles. As can be appreciated, the I3D ConvNet configurations exhibited a similar MAE increase along with the different quartiles. The results show that the first quartile achieved the best MAE during the experiments. These results may occur because of a minor timing variance during this quartile, and top-trained runners are quite consistent, while non-professional runners tend to perform less homogeneously. However, runners start to mix due to the distance running points and the different pace when the runners progress throughout the race; some runners may be in a recording point and other runners in a previous one. This increased the MAE to 32\% and 51\% for 2048-SB and 800-SB, respectively. Then, the MAE barely differs in the other quartiles until they reach 38\% and 62\% MAE degradation in Q4. Similarly, 3D ResNets (R3D) and C3D performance degrade regularly along with the different quartiles. Precisely, the results reported in Table \ref{tab:mae_exp} are from Q4. 

\section{Conclusion}
\label{sec:con}
This paper presented a novel approach to determine the CRT in a given recording point. The presented pipeline exploits ByteTrack to achieve tracking of the relevant subjects in the scene. This tracker allows to either remove the runner context (BB) or neutralize it by leaving the runner as the only moving object in the scene (SB). Then, I3D ConvNet is used as a feature extractor to feed several tested regression methods. In this regard, we have analyzed the results of the I3D ConvNet meticulously by splitting and combining the two 3D stream inputs -RGB and Flow-. The Multi-Layer Perceptron (MLP) reported the best results in every tested configuration. 

Contrary to classical HAR techniques, a higher number of I3D ConvNet embeddings -data from the penultimate inception block- provides a better result. Moreover, the output for each I3D ConvNet provides even a better result when they are concatenated instead of added. In addition, the reported experiments have demonstrated that context plays a key role during the classification process. The results show that accuracy regularly drops when the BB video clips are considered input to the classifiers.

Among the most relevant uses, it is evident that runner's performance monitoring must be mentioned. It can benefit from our proposal and, in general, any further achievements in the field by relieving the race staff from the need for tiring continuous attention for health issues. 



\section*{Acknowledgment}

This work is partially funded by the ULPGC under project ULPGC2018-08, the Spanish Ministry of Economy and Competitiveness (MINECO) under project RTI2018-093337-B-I00, the Spanish Ministry of Science and Innovation under project PID2019-107228RB-I00, and by the Gobierno de Canarias and FEDER funds under project ProID2020010024.



\bibliographystyle{IEEEtran}
%



\end{document}